\pdfoutput=1

\documentclass[11pt]{article}

\usepackage[]{acl}

\usepackage{times}
\usepackage{latexsym}
\usepackage{graphicx,xcolor}  
\usepackage{pxrubrica}        
\usepackage{url}
\usepackage{multirow}
\usepackage{color}
\usepackage{comment}
\usepackage{array}
\usepackage{booktabs}
\usepackage{amssymb}

\usepackage{CJKutf8}
\newcommand*{\ja}[1]{\begin{CJK}{UTF8}{ipxm}#1\end{CJK}}

\usepackage[T1]{fontenc}

\usepackage[utf8]{inputenc}

\usepackage{microtype}

\usepackage{inconsolata}

\title{Simplifying Translations for Children: \\ Iterative Simplification Considering Age of Acquisition with LLMs}

\author{Masashi Oshika\textsuperscript{1} \hspace{1ex}
  Makoto Morishita\textsuperscript{2} \hspace{1ex}
  Tsutomu Hirao\textsuperscript{2} \\
  {\bf Ryohei Sasano\textsuperscript{1} \hspace{1ex}
  Koichi Takeda\textsuperscript{1}} \\
    \textsuperscript{1} Graduate School of Informatics, Nagoya University\\
    \textsuperscript{2} NTT Communication Science Laboratories, NTT Corporation\\
    \texttt{oshika.masashi.f6@s.mail.nagoya-u.ac.jp}\\
    \texttt{\{makoto.morishita,tsutomu.hirao\}@ntt.com}\\
    \texttt{\{sasano,takedasu\}@i.nagoya-u.ac.jp} \\
  }
\begin{document}
\maketitle

\begin{abstract}
In recent years, neural machine translation (NMT) has been widely used in everyday life.
However, the current NMT lacks a mechanism to adjust the difficulty level of translations to match the user's language level.
Additionally, due to the bias in the training data for NMT, translations of simple source sentences are often produced with complex words.
In particular, this could pose a problem for children, who may not be able to understand the meaning of the translations correctly. 
In this study, we propose a method that replaces words with high Age of Acquisitions (AoA) in translations with simpler words to match the translations to the user's level.
We achieve this by using large language models (LLMs), providing a triple of a source sentence, a translation, and a target word to be replaced.
We create a benchmark dataset using back-translation on Simple English Wikipedia.
The experimental results obtained from the dataset show that our method effectively replaces high-AoA words with lower-AoA words and, moreover, can iteratively replace most of the high-AoA words while still maintaining high BLEU and COMET scores.
\end{abstract}

\section{Introduction}
\renewcommand{\thefootnote}{}
\footnote[0]{We have released our code and dataset at \url{https://github.com/nttcslab-nlp/SimplifyingMT_ACL24}}
\renewcommand{\thefootnote}{\arabic{footnote}}
Neural machine translation (NMT) has seen significant progress in recent years, making it practical for everyday use. 
As a result, more and more people are using it to meet their translation needs.
NMT systems can generate fluent and grammatically correct sentences in most cases. 
However, some people may have difficulty understanding the translations due to the complexity of the words used. 
This is an especially serious risk when children use NMT systems.
For example, when using an NMT system to translate textbooks used in the primary schools of another country, the system may use words that are not appropriate for primary school students in the country of the target language.
This is because NMT systems do not take into account the complexity of the words used to compose the translations.
Additionally, the parallel corpus used for the training, which mostly comes from the web, is biased because words used on the web are generally more complex than those used for children. 

Therefore, a mechanism is needed to control the complexity of words in NMT systems.
One way to measure such complexity is through the Age of Acquisition (AoA), which is the average age at which a person learns a particular word. Words with higher AoA are more difficult to learn. 
By controlling the AoA of words used in translations, we can simplify the translated sentence to a difficulty level that is more appropriate to the user. 
However, merely replacing words based on their AoA may not sufficiently simplify the sentence. Another approach is using paraphrase models trained with a parallel corpus of complex and simple sentences, but this method cannot control the AoA of words.

\begin{figure*}[t!]
\centering
\includegraphics[scale=0.32]{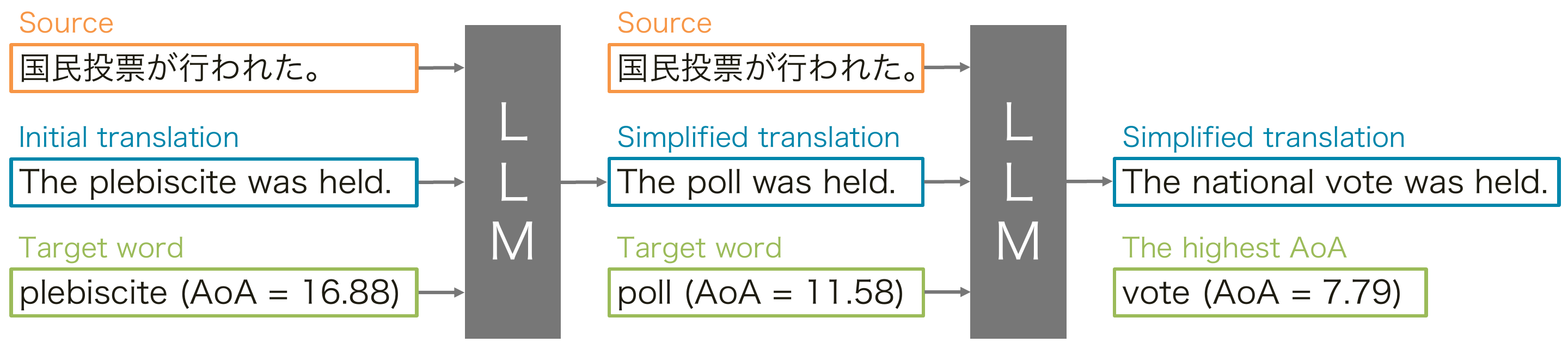}
\caption{Overview of the proposed method, which generates a simplified translation given the source sentence, initial translation, and target word. If the highest AoA in the output sentence is higher than the target age, the model will iteratively revise the sentence. The target word is defined as the word with the highest AoA in the initial translation.}
\label{fig:model}
\end{figure*}

This paper proposes a method for appropriately simplifying translations to particular user levels, such as children, through post-editing. 
Figure~\ref{fig:model} shows an overview of our method.
We replace a high-AoA word in a translation with a simple one by using a large language model (LLM) while giving the source sentence.
This technique allows us to replace not only the target word but also the surrounding words within the context of the sentence, which helps to simplify the overall sentence while preserving its original meaning.
Moreover, we can apply this process iteratively to simplify all high-AoA words in a translation or in a partially simplified sentence that still contains complex words.
Furthermore, users are able to customize the editing process on their own by specifying particular words that need to be replaced.
This feature allows a more personalized editing experience.

Since there is no adequate dataset available for assessing the proposed method, we created a benchmark dataset using a back-translation approach.
Initially, we translate a Simple English Wikipedia article into another language, and then we translate it back into English.
In this way, the back-translations are treated as translations of simple sentences in the second language. 
The original sentences of the Simple English Wikipedia article serve as the reference simple sentences, and their corresponding back-translations serve as the sentences that need to be simplified.

Based on the results obtained using our benchmark dataset, we confirmed that our method outperforms the baseline methods, namely, MUSS~\cite{muss}, a simple post-editing approach, AoA-constrained NMT, an automatic post-editing approach, and LLM-based translation. 
Our method achieved the highest BLEU score while also maintaining the quality of simplification. Specifically, our method successfully replaced words with high AoA with those having lower AoA.

The contributions of this paper are as follows:
\begin{enumerate}
    \item 
    We automatically created a dataset based on Age of Acquisition (AoA) for text simplification from a monolingual corpus.
    \item 
    We propose a simplification technique using large language models (LLMs) to iteratively replace high-AoA words with lower-AoA ones. Furthermore, we demonstrate that our proposed method outperforms the baseline methods in both translation and simplification, as shown by the experimental results. 
\end{enumerate}

\section{Related Work}

\subsection{Text Simplification}
Text simplification is the task of transforming complex sentences into simple sentences that use basic vocabulary and grammar.
This task shares the same objective as ours, which is to produce simplified sentences for children and beginner learners of a second or other language.
The most common methods used to accomplish the text simplification task are adapting statistical machine translation~\cite{wubben,xu} and applying neural-based sequence-to-sequence models~\cite{zhang,guo} trained with a parallel corpus consisting of complex and simple sentences~\cite{coster}.
However, these methods cannot be controlled for specific simplification levels, including AoA. 
Our method can handle the more detailed aspects of simplification by iterating revisions with LLMs, permitting a more personalized simplification that would prove difficult for the previous methods.

\subsection{Automatic Post-Editing}
Automatic post-editing is a technique widely used to correct or improve machine translation outputs.
Recently, transformer-based post-editing models have been widely used for this task~\cite{bhattacharyya}.
In particular, methods that iteratively edit machine translation outputs have been proposed in recent years~\cite{gupta,chen2023iterative}.
\newcite{gupta} proposed an iterative editing method using special tokens for keeping, deletion, controlling sentence length, and paraphrasing.
\newcite{chen2023iterative} used source sentences and mistranslated examples as prompts and then made iterative post-edits with LLMs.
These methods are similar to ours, which edits machine translation outputs.
However, automatic post-editing conventionally aims to correct errors in generated sentences to improve translation quality.
In contrast, our work does not focus on correcting errors but attempts to control sentence difficulty, which is an important but challenging task.

\subsection{Complexity-control Method}
There are several methods to generate sentences while controlling their difficulty.
\newcite{agrawal,tani-etal} proposed multi-level complexity controlling machine translation (MLCCMT) methods, which can control the difficulty of generated sentences at multiple levels.
In addition, several methods have been proposed in the field of text simplification~\cite{scarton,nishihara,chi,agrawal-emnlp}.
However, most of these methods define difficulty in sentence-level granularity, and thus they do not control difficulty at the level of individual words.
In contrast, we achieve word-level control in our simplification by specifying the target word.
In addition, since our method performs word-level simplification, each user can specify the word that he or she does not understand, which allows interactive adjustment of the sentence's difficulty level to match individual user's needs.

\section{Dataset Creation}
\subsection{Procedure using Back-Translation}
\begin{figure}[t!]
\centering
\includegraphics[scale=0.27]{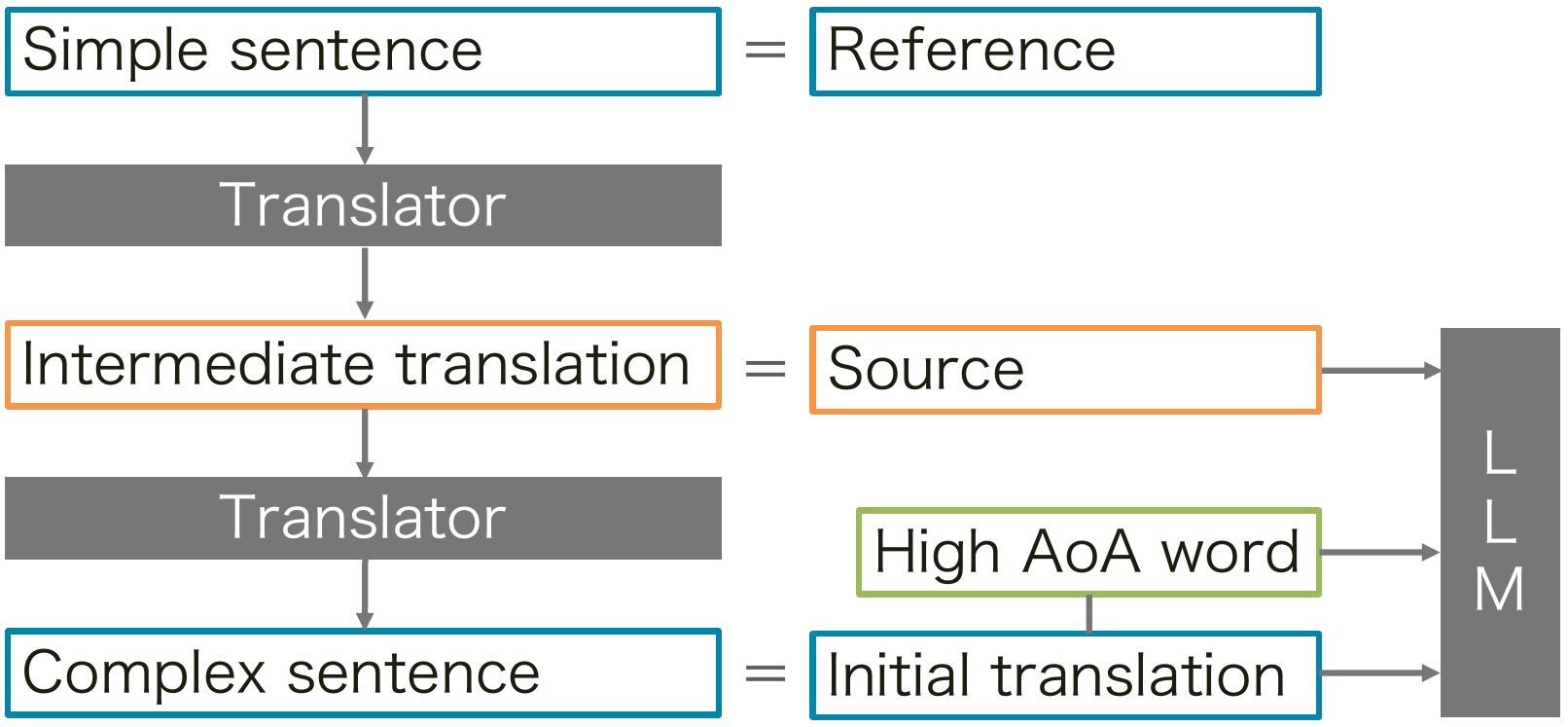}
\caption{Procedure for creating a data set and the corresponding input to the LLM.}
\label{fig:data}
\end{figure}

\label{sec:data}
In order to evaluate our simplification technique for machine translation systems with a focus on Age of Acquisition (AoA), we need a dataset of translations that contain words with high AoA.
We also need corresponding reference sentences that only include appropriate AoA words. Unfortunately, datasets of such pairs of sentences are not publicly available. 
Therefore, we created a dataset by automatically paraphrasing a monolingual corpus.
Figure~\ref{fig:data} shows the procedure for creating this dataset.
To obtain translations including high-AoA words automatically, we use a back-translation approach, which translates a reference, a simple sentence, into another language (an intermediate translation) and then back into the original language using machine translation models~\cite{sennrich,edunov,zhang_bt}.
Current machine translation systems often produce translations with high-AoA words due to biases in a training parallel corpora, even when the original sentences are intended for beginners.
This means that when translating simple English sentences into another language, the translations often contain complex words in the target language.
As a result, back-translations from the second language may end up having more complex words than the original sentences from which they were derived.
Note that our method has the advantage of not requiring a simplification corpus that consists of complex-simple sentence pairs.
Therefore, this method can be used for any language with sufficient monolingual data and an AoA list\footnote{For example, AoA for Spanish words was estimated by \newcite{Alonso2014SubjectiveAN}.}.

When training or testing models to simplify translations for children of a certain age, we choose pairs of source sentences that contain words with an AoA greater than that age and the corresponding reference sentences that contain words with an AoA less than that age.

\subsection{Creating Dataset from Simple English Wikipedia}

We used sentences from Simple English Wikipedia\footnote{\url{https://huggingface.co/datasets/wikipedia/viewer/20220301.simple}} as a reference in our experiment. 
This was because it is available to the public and written in easy English for beginners. As part of the pre-processing step, we excluded the titles and section titles of entries. 
Consequently, 1,754,964 sentences were extracted.

We then performed back-translation on the dataset.
We applied an English-to-Japanese translator to Simple English Wikipedia articles.
After that, we applied a Japanese-to-English translator to the translated Japanese articles to obtain alternatives to the original English articles.
Finally, we selected pairs of English sentences that showed a difference greater than 0.5 between the words with highest AoA in the reference and source sentences, respectively.

\begin{figure}[!t]
\centering
\includegraphics[width=\linewidth]{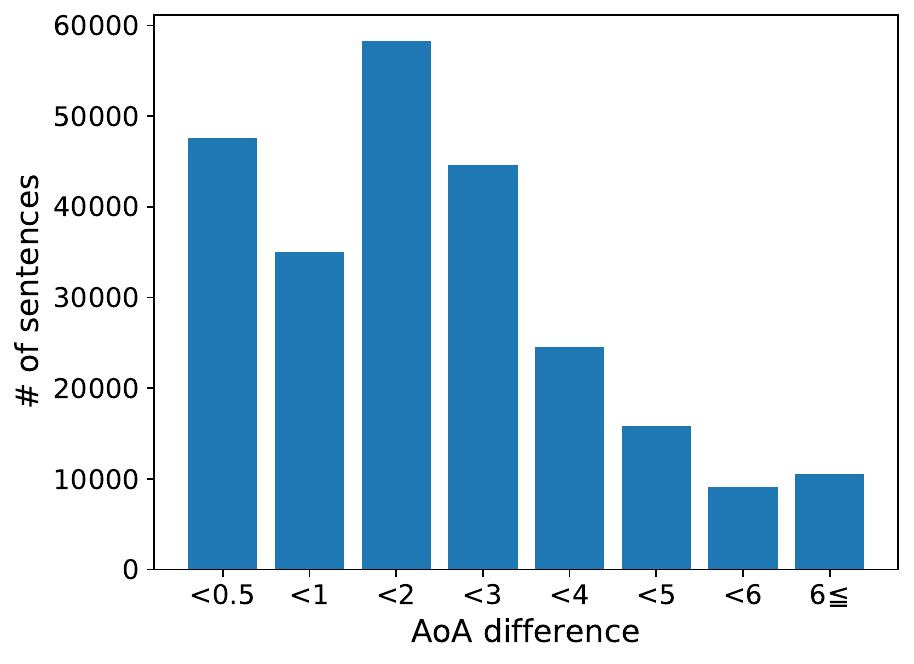}
\caption{Distribution of the AoA difference in the created dataset, showing only the sentences where the AoA difference is greater than 0. AoA difference was 0 for 1,164,870 sentence pairs for 66\% of the dataset.}
\label{fig:sta}
\end{figure}

In Figure~\ref{fig:sta}, we show the distribution of the differences between the highest AoA of the reference and that of the corresponding back-translation.
From the figure, we found that the highest AoA of the words in the back-translation is sometimes greater than those in the original reference sentences.
This finding appears to support our motivation, i.e., the need for a mechanism to replace complex words with simple ones for children.

After applying the filter, 235,331 sentences remained, which we divided into three sets: training, development, and test, at a ratio of 8:1:1.
To provide back-translation, we trained two MT systems, English-to-Japanese and Japanese-to-English, using a transformer-based encoder-decoder model.
Our training used JParaCrawl v3.0~\cite{morishita-etal-2022-jparacrawl}, the largest parallel corpus for English-to-Japanese/Japanese-to-English translation.
Accordingly, our back-translation achieved a BLEU score of 48.3 and a COMET score of 90.0, indicating sufficient translation performance.

\section{Proposed Method}
Post-editing of initial translations is a promising approach to providing child-appropriate translations, given the difficulty of controlling the AoA of words in MT systems.
One approach to achieve this is by simply replacing high AoA words in translations with lower AoA ones using a thesaurus.
However, word-to-word replacement may sometimes change the meaning of the original sentences due to mismatched collocations. On the other hand, existing text simplification techniques are also available, but they do not guarantee that simplified translations consist of only appropriate AoA words.

To address these issues, we propose a method for iteratively applying LLMs to explicitly replace a high-AoA word with a lower one.
LLMs are advanced language models that can reconstruct phrases containing high-AoA words, meaning they can significantly alter translations without changing their meanings.

We fine-tuned an LLM to generate a simplified sentence from a given initial translation, the source sentence\footnote{
The source sentence is the intermediate translation of the reference sentence, as shown in Figure~\ref{fig:data}.},
and a word whose AoA is greater than the specific value in the translation (Figure~\ref{fig:model}).
We assumed that using the source sentences to simplify the translations would maintain the original meaning despite replacing the high-AoA words.
High-AoA words in translations are enclosed in \texttt{<edit>} tags.
Since our method determines the words to be edited based on their AoA, 
it allows for easy tailoring of the MT system based on the user's age.
In other words, 
when the MT system users are $n$-year-old children, our system generates translations by replacing words with an AoA greater than $n$ with words with an AoA less than or equal to $n$.
In addition, since users themselves can specify words, it is possible to simplify a sentence by specifying words that a particular user does not understand, regardless of their age.

This method can be used iteratively, and thus if a translation contains multiple words that need to be replaced, all of these words can be replaced by repeatedly specifying words.
In addition, if the AoA does not decrease after one round of simplification, it is possible to iteratively continue simplification until the AoA is below the criterion by again specifying the target words.
The prompts used in the experiments are shown in Table~\ref{tb:prompt}.

\begin{table*}[!ht]
\centering
\small
\begin{tabular}{ll}
\toprule
\!\!\multirow{6}{*}{\begin{tabular}[c]{@{}l@{}}Proposed \\ method\hspace{-3pt} \end{tabular}} 
&\!\!Instruction: Translate the following source language sentence based on the machine translation by simplifying \\ 
&\!\!the words surrounded by \texttt{<edit>}.\\
&\!\!\#\#\# source language sentence: \scalebox{0.8}{\ja{この用語は、アルバム上の特定の曲を数字で表すためによく使用されます。}}\\
&\!\!\#\#\# machine-translation: This term is often used to \texttt{<edit>}denote\texttt{<edit>} certain songs on the album \\ &\!\!by numbers. \\
&\!\!\#\#\# translation:\\
\midrule
\!\!\multirow{4}{*}{\begin{tabular}[c]{@{}l@{}}Direct \\ Translation\hspace{-3pt} \end{tabular}} 
&\!\!You are a Japanese-English translator who only generates words that ten-year-old children can understand.\\ 
&\!\!Output the translation only.\\
&\!\!\#\#\# Source Japanese \scalebox{0.8}{\ja{この用語は、アルバム上の特定の曲を数字で表すためによく使用されます。}}\\
&\!\!\#\#\# Translated English:\\
\midrule
\!\!\multirow{6}{*}{\begin{tabular}[c]{@{}l@{}}Multi- \\ word\hspace{-3pt} \end{tabular}} 
&\!\!Instruction: Translate the following source language sentence based on the machine translation by simplifying\\
&\!\!the words surrounded by \texttt{<edit>}.\\
&\!\!\#\#\# source language sentence: \scalebox{0.8}{\ja{この用語は、アルバム上の特定の曲を数字で表すためによく使用されます。}}\\
&\!\!\#\#\# machine-translation: This term is often used to \texttt{<edit>}denote\texttt{<edit>} \texttt{<edit>}certain\texttt{<edit>} songs on the\\
&\!\!album by numbers. \\
&\!\!\#\#\# translation:\\
\bottomrule
\end{tabular}
\caption{Prompts used in experiments. The original prompts are in Japanese, but here English-translated prompts are displayed for readability.}
\label{tb:prompt}
\end{table*}

\section{Experiments}
\subsection{Settings}
As an LLM, we used the pre-trained English-Japanese bilingual GPT-NeoX, which consists of 3.8 billion parameters, provided by Rinna\footnote{\url{https://huggingface.co/rinna/bilingual-gpt-neox-4b-instruction-sft}}.
To adapt the model to our task, we fine-tuned it using LoRA~\cite{lora}.
LoRA was applied to all linear layers in the model, resulting in 25,952,256 trainable parameters, which is 0.68\% of all model parameters.
The hyperparameters of LoRA were set to $r=16$ and $\alpha=32$.
The learning rate for fine-tuning was linearly decayed, with an initial rate of 1$e$-5.
We set the batch size to 16 and used AdamW as the optimization method.
Fine-tuning was performed for 10 epochs, and the model with the smallest loss in the dev set was used to generate the test set.

In this study, the target age was set to 10 years old.
In other words, the test set consists of data in which the highest AoA in the back-translation to be simplified was greater than 10, and the highest AoA in the reference sentences was less than 10.
The total number of sentences in the test set was 8,289.

\subsection{Evaluation Metrics}
To evaluate how well the model succeeds in generating a simplified translation, we need to evaluate it from two viewpoints: machine translation accuracy and simplification quality.
Machine translation accuracy indicates how fully the source sentence's meaning is maintained in the simplified sentence, while simplification quality indicates how well the model generates easy sentences.
As machine translation metrics, we employed BLEU~\cite{bleu}, which evaluates text based on n-gram agreement, and COMET~\cite{comet}, which evaluates text based on embedded similarity.
For the COMET model, we used \texttt{Unbabel/wmt22-comet-da}\footnote{\url{https://huggingface.co/Unbabel/wmt22-comet-da}}.
As the text simplification metric, we used SARI~\cite{xu}, which is evaluated based on the number of paraphrases and other factors from a triple of source, reference, and generated sentences.
We also used FKGL~\cite{fkgl}, which is evaluated based on the number of words per sentence and syllables per word in the sentence.
In addition, we employed Dale-Chall Readability~\cite{dale}, which assesses the readability of English text based on the average sentence length and the percentage of difficult words not found in a list of pre-defined common words.
The Average AoA is the calculated average of the highest AoA for each sentence generated by each method.
The success rate of simplification is the percentage of sentences in which the highest AoA in the sentence is lower than the target age.


\renewcommand{\arraystretch}{1.1}
{
\begin{table*}[!t]
\centering
\small
\begin{tabular}{lccc}
\toprule
Method & Initial translation & Source language sentence & Target word \\\midrule
MUSS & $\checkmark$ & $\times$ & $\times$ \\
Constraint Generation & $\times$ & $\checkmark$ & $\times$\\
APE & $\checkmark$ & $\checkmark$ & $\times$ \\
Direct Translation& $\times$ & $\checkmark$ & $\times$ \\
Multi-Word & $\checkmark$ & $\checkmark$ & $\checkmark$ \\
Proposed mmethod & $\checkmark$ & $\checkmark$ & $\checkmark$ \\
\bottomrule
\end{tabular}
\caption{Input information used by each baseline method and proposed method in generating sentences.}
\label{tb:input}
\end{table*}
}

\renewcommand{\arraystretch}{1.1}
{\tabcolsep = 4.5pt
\begin{table*}[!ht]
\small
\centering
\begin{tabular}{lcccccccccccc}
\toprule
& Initial      &      & Constraint &    & Direct  &\multicolumn{2}{c}{Multi-word} &  \multicolumn{5}{c}{Proposed method} \\
& translation  & MUSS & Generation & APE& Translation & 1 & 5 & 1 & 2 & 3 & 4 & 5 \\
 \midrule
\begin{tabular}[c]{@{}c@{}}\# of generated \\[-2pt] sentence\end{tabular} & 8,289 & 8,289  & 8,289 & 8,289 &  8,289 &  8,289 & 601  & 8,289  & 1,232  & 599  & 387  & 287  \\
BLEU↑      & 38.4  & 26.9  & 39.6 / 39.7   & 44.8 & 30.5 & 44.3   & 44.6 & 44.9   & 45.0  &  45.0 & 45.0  & \textbf{45.1}    \\
COMET↑     & 87.3 & 83.4 & 86.4 / 86.2  & 87.9 & 84.8 &  \textbf{88.3} & 88.2 &  88.2 & 88.2  & 88.2 &  88.2 & 88.2   \\
SARI↑      & 53.2 & 42.6 & 53.9 / 53.5  & 58.6 & 49.5  &59.3  & 59.8 & 59.8  & 60.0 &\textbf{60.1}& \textbf{60.1} & \textbf{60.1}  \\
FKGL↓     & 9.26  & \textbf{6.49}  &  8.75 / 8.66  & 8.78 & 8.37 & 8.85 & 8.78 & 8.69 & 8.65  & 8.64  & 8.64  & 8.64   \\
Dale-Chall↓ & 10.0 &  8.82 &  9.59 / 9.55  &  9.70 &  \textbf{8.17} & 9.58  & 9.52 & 9.50 & 9.47  & 9.46   & 9.46  & 9.45 \\
Average AoA & 11.58 & 9.18  &  8.86 / 8.62  & 8.87 & 9.14 & 8.76  & 8.23 &8.42  & 8.18 &  8.09 &  8.05 & 8.03   \\
Success Rate↑ & 0.0 & 0.62 & 0.82 / 0.89 & 0.71 & 0.66 & 0.78  &  0.94 & 0.85  &  0.93 & 0.95 &  \textbf{0.97} & \textbf{0.97} \\
\bottomrule
\end{tabular}
\caption{Experimental results.
The second row of the ``Multi-word'' and ``Proposed method'' columns indicate the number of iterations.
The scores on the left side were obtained by the beam size of 6, while those on the right side were obtained by the beam size of 20 in the ``Constraint Generation'' column.
``\# of generated sentences'' indicates the number of sentences generated in each iteration. If the highest AoA in the generated sentence is less than 10 (target age), the sentence will not be included in the next iteration for Multi-word and Proposed methods. 
During evaluation, the entire test set (8,289 sentences) is used, not just the generated sentences.}
\label{tb:result}
\end{table*}
}

\renewcommand{\arraystretch}{1}

\subsection{Compared Methods}
Here, we compare our method with the following baseline methods:\\
\noindent\textbf{MUSS} \cite{muss} is an unsupervised simplification model based on BART and trained with a parallel dataset for paraphrasing.
To obtain simplified translations, we apply MUSS to initial translations, i.e., we used it as a post-editor.\\
\noindent\textbf{Constraint Generation} is a translation model that 
restricts the generation of words whose AoA is more than ten.
Specifically, if a hypothesis contains a word with an AoA of ten or more at each generation time-step, the score of the hypothesis is set to $-\infty$.
The beam size for search is set to 6 and 20.
This method does not output anything if the translation failed in the restricted vocabulary.
Therefore, if the translation fails, the initial translation in the test set is treated as the generated sentences.
Note that Constraint Generation is incorporated in a machine translation model, i.e., it is not a post-editing approach.
This method uses the same neural machine translation model that was used for dataset creation~\cite{morishita-etal-2022-jparacrawl}.
\\
\noindent\textbf{APE} is an automatic post-editing model trained on our dataset. 
As the APE model, we trained the transformer-big model that generates the reference simple translation given a pair of source language sentences and initial translations, which are concatenated with a special token (i.e., <SEP>).
The model architecture is the same as the original Transformer, and it is trained with a cross-entropy loss.
This method is one of the simple but strong baselines in the APE task, as seen by participants in the recent WMT APE task \cite{bhattacharyya} also employing similar Transformer-based approaches.
\\
\noindent\textbf{Direct Translation} is an LLM-based translation model that directly translates source language sentences into simple target sentences.
We utilized GPT-3.5-turbo as the LLM and provided it with the prompt shown in the second row of Table \ref{tb:prompt}.\\
\noindent\textbf{Multi-word} is a variant of our proposed method. 
We provide LLMs with all words having an AoA above ten in a translation within a single iteration, instead of providing words iteratively.
The prompt for this model is shown in the third row of Table \ref{tb:prompt}.\\
The input data utilized for each method is shown in Table \ref{tb:input}.

\section{Results and Discussion}
\subsection{Main Results}
Table~\ref{tb:result} shows the experimental results.
Although MUSS achieved the best FKGL and Dale-Chall scores, it scored the worst in BLEU and COMET, indicating that MUSS simplifies translations, but this simplification leads to inaccuracies.
In addition, the low success rate implies that this level of simplification does not align with our goal of creating simplified translations for ten-year-old children.

Constraint Generation yielded better results in all metrics except COMET than the initial translation for both beam sizes by effectively suppressing the generation of words with an AoA greater than 10.
When comparing beam sizes 6 and 20, the latter beam size generally yielded better results, with the exception of COMET.
However, its success rate was only 0.89 with the beam size of 20, indicating that Constraint Generation often failed to generate words with an AoA of less than 10.
It was also unsuccessful in generating translations for about 13\% of the sentences due to the limitations of beam search, such as the repetition problem.
These findings suggest the limitations of constraint generation with simple beam search, making it difficult to satisfy the constraints during decoding.

On the other hand, we found that APE produced better results than Constraint Generation. 
However, the Success Rate was the third worst, which suggests that it was challenging to replace high-AoA words.
In addition, this also suggests that simplification might be achieved by ignoring high-AoA words. We also tested APE without source sentences (namely, intermediate translations) but did not observe any improvement in its performance compared to APE with intermediate translations.

The direct translation method improved the FKGL and Dale-Chall scores, but it led to a notable decrease in both BLEU and COMET scores. 
Additionally, the success rate of the method is only 0.66. These findings suggest that direct translation, which involves translating the source language sentence into a simple target sentence, cannot fully ensure the accuracy and complexity of translations.

After the first round of iterations, the proposed method showed improvements in all metrics when compared to the initial translations. This indicates that our method can simplify a translation while still retaining the original meaning. In addition, our method achieved significantly better scores than MUSS, Constraint Generation, APE, and Direct Translation, with the exceptions being FKGL and Dale-Chall. 

\begin{table*}[!ht]
\centering
\small
\begin{tabular}{ll}
\toprule
\multirow{4}{*}{\begin{tabular}[c]{@{}l@{}}w/o \\ intermediate\hspace{-3pt} \end{tabular}} 
&\!\!Instruction: Simplify the following machine translation by simplifying the words surrounded by \texttt{<edit>}.\\
&\!\!\#\#\# machine-translation: This term is often used to \texttt{<edit>}denote\texttt{<edit>} certain songs on the album \\
&\!\!by numbers.\\
&\!\!\#\#\# simplified sentence:\\
\midrule
\multirow{4}{*}{\begin{tabular}[c]{@{}l@{}}w/o \\ word\end{tabular}} 
&\!\!Instruction: Translate the following source language sentence based on the machine translation.\\
&\!\!\#\#\# source language sentence: \scalebox{0.8}{\ja{この用語は、アルバム上の特定の曲を数字で表すためによく使用されます。}}\\
&\!\!\#\#\# machine-translation: This term is often used to denote certain songs on the album by numbers.\\
&\!\!\#\#\# translation:\\
\midrule
\multirow{3}{*}{\begin{tabular}[c]{@{}l@{}}w/o \\ intermediate\hspace{-3pt} \\ and word\end{tabular}} 
&\!\!Instruction: Simplify the following machine translation.\\
&\!\!\#\#\# machine-translation: This term is often used to denote certain songs on the album by numbers.\\
&\!\!\#\#\# simplified sentence: \\
\bottomrule
\end{tabular}
\caption{Prompts used in ablation study.}
\label{tb:prompt2}
\end{table*}

\renewcommand{\arraystretch}{1.1}
\begin{table*}[!t]
\small
\centering
\begin{tabular}{lccccccWc{1.5cm}Wc{0.5cm}}
\toprule
 & \multicolumn{2}{c}{Proposed method}&  \multicolumn{2}{c}{w/o intermediate} &  \multicolumn{2}{c}{w/o word} &  \multicolumn{2}{c}{w/o intermediate and word} \\
 & 1 & 5 & 1 & 5 & 1 & 5 & 1 & 5 \\\midrule
\# of generated sentences &8,289 &287 &8,289 &326&8,289 &779
& 8,289 & 585\\
BLEU↑        & 44.9 & \textbf{45.1} & 40.0 & 39.4  &44.6&45.0 & 40.2 & 39.4   \\
COMET↑       & 88.2 & 88.2 & 86.9 & 86.4 & \textbf{88.3} & \textbf{88.3} & 87.0 & 86.5 \\
SARI↑        & 59.8 & \textbf{60.1} & 55.5 & 55.4& 59.3 &60.0 & 55.5 & 55.4  \\
FKGL↓        & 8.69 & 8.64 & 8.67 & 8.60 &8.60 & 8.58 & 8.58 & \textbf{8.46}  \\
Dale-Chall↓ &  9.50 & 9.45  & 9.51   & 9.44  &  9.59 & 9.53  & 9.48 & \textbf{9.39} \\
Average AoA  & 8.42 & 8.03 & 8.46 & 7.99  &8.43& 8.17 & 8.51 & 8.00  \\
Success Rate↑& 0.85 & \textbf{0.97} & 0.83 & \textbf{0.97} &0.74 & 0.91 & 0.80 &0.94\\
\bottomrule
\end{tabular}
\caption{Experimental results of ablation study. Numbers under the method names indicate the number of iterations.}
\label{tb:result_a}
\end{table*}
\renewcommand{\arraystretch}{1}

With further iterations, our method showed slight improvements in BLEU and COMET but significant improvements in Average AoA and Success Rate. These results clearly demonstrate the effectiveness of our iterative simplification approach.

When comparing one-word replacement per iteration with multi-word replacement per iteration, it was found that the former achieved a better Success Rate, while both approaches obtained similar BLEU and COMET scores. 
On further analysis, it was observed that after five iterations, the one-word replacement approach outperformed multi-word replacement on BLEU and Success Rate. 
Another interesting finding was that the Average AoA was lower in one-word replacement than in multi-word replacement. 
These results suggest that multi-word replacement is a more challenging task for LLMs than one-word replacement.

\begin{figure}[!t]
\centering
\includegraphics[width=\linewidth]{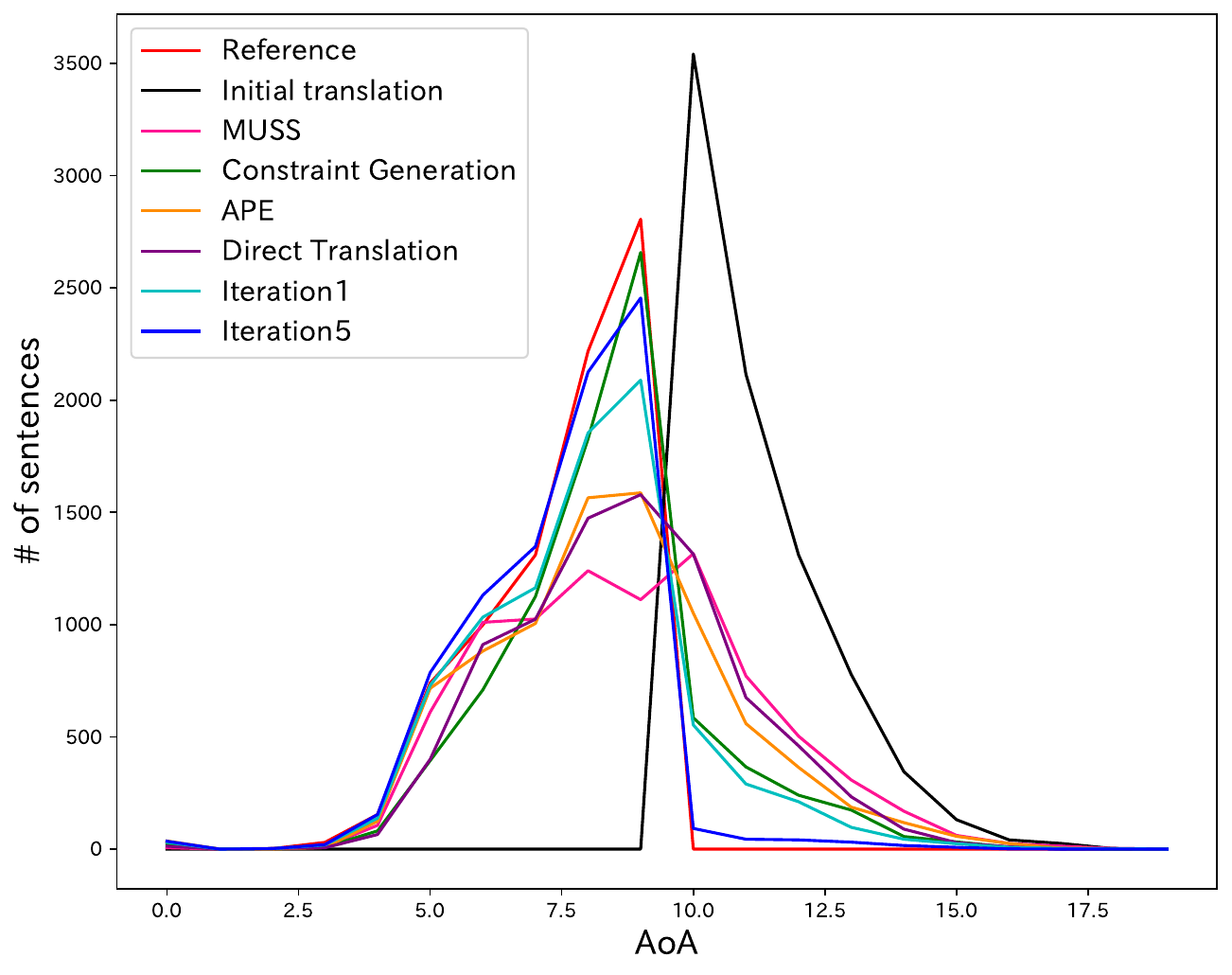}
\caption{Statistical plots of highest AoA of the generated sentence by each method.}
\label{fig:graph}
\end{figure}

Figure~\ref{fig:graph} shows the distribution of the highest AoA for each sentence and for different methods.
The red line represents the reference, while the blue line represents the initial translation. After the first round of iterations, our method generated words with an AoA of 10 or higher. However, with further iterations, these words were replaced with low-AoA words. Although the Constrained Generation method and APE use words with low AoA, in some cases, they fail to generate low AoA sentences. As a result, sentences with an AoA of 10 or more are generated. Furthermore, MUSS generates a large number of words with an AoA of 10 or higher.

In short, our method simplifies given translations for a certain age group of children without reducing translation performance, as evidenced by these findings.

\subsection{Ablation Study}
In order to demonstrate the effectiveness of the proposed method, we conducted ablation studies using three different models. 
The first model, called ``w/o intermediate,'' only provides translations and target words to LLMs using prompts.
The prompts used in this model are shown in the first row of Table~\ref{tb:prompt2}. 
The second model, called ``w/o word,'' only provides translations and intermediate translations to LLMs, and the prompts this model uses are shown in the second row of Table~\ref{tb:prompt2}.
The third model, called ``w/o intermediate and word,'' only provides translations to LLMs, and the prompts it uses in this model are shown in the third row of Table~\ref{tb:prompt2}.

Table~\ref{tb:result_a} exhibits the results obtained from the first and fifth iterations of each model.
It is evident from the table that when intermediate translations were excluded, the translation performance showed a significant degradation.
This is particularly clear in the results obtained from the ``w/o intermediate'' and ``w/o intermediate and word'' columns. The findings suggest that the input of intermediate translations is necessary to preserve the meaning of translations while simplifying them. 
Additionally, excluding target words resulted in a decrease in Success Rates, which is evident from the ``w/o word'' and ``w/o intermediate and word'' columns. 
These results suggest that specifying words for editing is essential for good simplification with fewer iterations.
In summary, the presence of intermediate translations and that of target words are key indicators of successful simplification.

\begin{table*}[!t]
\centering
\small
\begin{tabular}{lll}
\toprule
& Sentence & Highest AoA word\\\midrule
Source & \scalebox{0.9}{\ja{この用語は、アルバム上の特定の曲を数字で表すためによく使用されます。}} & \\\midrule
\begin{tabular}[c]{@{}l@{}}Initial \\ translation\end{tabular} & This term is often used to denote certain songs on the album by numbers. & denote (11.24) \\\midrule
MUSS & This term is often used to show how many songs are on the album. & term (8.28)\\\midrule
\begin{tabular}[c]{@{}l@{}}Constraint \\ Generation\end{tabular} & The term is often used to describe a specific song on an album in numbers. & specific (9.28) \\\midrule
APE & The term is often used to describe certain songs on the album by numbers. & term (8.28) \\ \midrule
\begin{tabular}[c]{@{}l@{}}Direct \\ Translation\end{tabular} & This word is often used to represent a specific song on an album with numbers. &represent (10.33) \\ \midrule
Multi-word & The term is often used to describe certain songs on an album by numbers. & term (8.28) \\ \midrule
Iteration 1 & The term is often used to represent a particular song on a given album by numbers. & represent (10.33) \\\midrule
Iteration 2 & The term is often used to describe certain songs on a given album by numbers. & term (8.28) \\\midrule
Reference & The term is often used to mean a specific song on the album by number. & specific (9.28) \\\bottomrule
\end{tabular}
\caption{Successful generation example. Proposed method could iteratively decrease the AoA to below the target age.}
\label{tb:gene1}
\end{table*}

\begin{table*}[!t]
\centering
\small
\begin{tabular*}{\linewidth}{lll}
\toprule
& Sentence & \!\!Highest AoA word\\\midrule
Source & \scalebox{0.9}{\ja{しかし、その起源は、453年後の1951年に外国人によって最初に調査されました。}} & \\\midrule
\begin{tabular}[c]{@{}l@{}}Initial \\ translation\end{tabular} & But its origin was first investigated by foreigners in 1951, 453 years later. & \!\!foreigners (10.39) \\\midrule
MUSS & But foreigners first looked at its origin in 1951, 453 years later. & \!\!foreigners (10.39)\\\midrule
\begin{tabular}[c]{@{}l@{}}Constraint \\ Generation\end{tabular} & However, its roots were first investigated by a foreign citizen in 1951, 453 years later. & \!\!investigated (9.0) \\\midrule
APE & However, its origin was first investigated by foreign people in 1951, 453 years later. & \!\!investigated (9.0) \\\midrule
\begin{tabular}[c]{@{}l@{}}Direct \\ Translation\end{tabular} & But, its origin was first investigated by foreigners in 1951, 453 years later.
 &\!\!foreigners (10.39) \\ \midrule
Multi-word & Its roots, however, were first explored by outsiders in 1951, 453 years later. & \!\!outsiders(9.75)\\\midrule
Iteration 1 & But its origin was first explored by foreigners in 1951 after 453 years. & \!\!foreigners (10.39) \\\midrule
Iteration 2 & However, its origins were first investigated by foreign people in 1951 after 453 years. & \!\!origins (10.25) \\\midrule
Iteration 3 & However, its origins were first explored in 1951 by foreigners 453 years later. & \!\!foreigners (10.39) \\\midrule
Iteration 4 & But its origins were first examined in 1951 by foreign people 453 years later. & \!\!origins (10.25) \\\midrule
Iteration 5 & Its origins, however, were first looked at in 1951 by foreign researchers, after 453 years. & \!\!origins (10.25) \\\midrule
Reference & Its source, however, was first explored by non-native people in 1951, 453 years later.& \!\!native (9.20) \\\bottomrule
\end{tabular*}
\caption{Example of erroneous generation. Proposed method generated ``foreigners'' and ``origins,'' in which AoA is not less than the target age.}
\label{tb:gene2}
\end{table*}

\subsection{Examples of Simplification}
\subsubsection*{Successful Simplification}

In Table~\ref{tb:gene1}, we can see an example of successful simplifications in terms of AoA. In MUSS, the high-AoA word ``denote'' was replaced with a lower one, ``show,'' resulting in all words having an AoA of less than 10.
However, this change altered the meaning of the original translation.
Constraint Decoding and APE were able to replace ``denote'' with ``describe'' while maintaining an AoA of less than 10 for all words.
Our method, on the other hand, was initially unable to replace ``denote'' with a lower AoA word, but with more iterations, it eventually succeeded in doing this while keeping all words at an AoA of less than 10.

\subsubsection*{Erroneous Simplification}\label{sec:erroneous}
Table~\ref{tb:gene2} shows an example of erroneous generation.
After the second iteration, our approach was successful in replacing the high-AoA word ``foreigners'' with a simpler phrase, ``foreign people.'' However, the word ``origins'' remains another complex word. Unfortunately, in the subsequent attempt, the word ``foreigners'' was generated again. This indicates that our method sometimes fails to simplify translations even after five iterations. To overcome this limitation, providing previous simplified sentences in each iteration could be an effective solution.

\section{Conclusion}
We proposed a method to provide easy-to-understand translations for children.
Our method involves iteratively replacing high-AoA words in a translation with lower-AoA ones using LLMs.
This is done by providing a triple of the source sentence, the translation, and the target word to be replaced.
We also automatically generated a dataset from a monolingual corpus to evaluate simplification based on the back-translation technique.
Moreover, we compared our method with baseline methods, namely, MUSS, Constraint Generation, Automatic Post-Editing, and Direct Translation, and we tested them on ten-year-old children (AoA=10).
The results show that our method successfully replaced high-AoA words with lower-AoA ones while maintaining the highest BLEU and COMET.
In particular, our method simplified 97\% of complex translations after five iterations. Furthermore, ablation studies identified the best choice as the approach of replacing one complex word with a simpler one in each iteration.

\section*{Limitations}
One serious limitation of the proposed method is its computational cost.
In our experiments, our proposed method required an average of 0.5 seconds to generate a sentence in each iteration.
However, note that this is not the slowest speed among the compared methods.
The constraint generation method, which is the slowest, requires an average of 1.8 seconds to generate a sentence, since it requires checking whether the hypothesis includes high-AoA words in each time step.
We assume the computation could be faster by distilling the LLMs or utilizing smaller LLMs.

Another limitation is that our method sometimes fails to simplify the sentence to meet the target age, as described in Section~\ref{sec:erroneous}.
In that case, the model became stuck in the loop of two high-AoA words, `foreigners' and `origins.'
We hypothesized that this is because the model does not know the history of the previous iterations, and thus the results could be improved by feeding this history to the model.

We also plan to conduct extensive experiments with other experimental settings.
For example, This study carried out experiments with a target age of 10, but it would be interesting to see the effect of varying the target age.
This paper focused on simplifying English, but we plan to extend it to other languages, such as Spanish, which already has an AoA estimation~\cite{Alonso2014SubjectiveAN}.
We believe we can extend this method to a language with no AoA list by estimating the word complexity using unigram probability in the corpus, since a rare unigram would be difficult while and a frequently used one would be an easy word.
However, we left these pursuits for future work.

\bibliography{custom}

\appendix

\end{document}